\documentclass[aac]{iosart2x}
\usepackage{float}
\usepackage{color}
\usepackage{xcolor}
\usepackage{picture}
\usepackage{listings}
\usepackage{caption}
\usepackage{makeidx}
\usepackage{graphicx}
\usepackage{amsmath}
\usepackage[english]{babel}
\usepackage{commath}
\usepackage{amssymb}
\usepackage{subcaption}
\usepackage[utf8]{inputenc}
\usepackage{hyperref}
\usepackage[nameinlink,noabbrev]{cleveref}
\usepackage{xfrac}
\usepackage{booktabs}
\usepackage[T1]{fontenc}
\usepackage{enumitem} 

\hypersetup{%
  colorlinks=true,
  linkcolor=blue,
  linkbordercolor=red,
  citecolor=blue,
  linkcolor = blue,
  anchorcolor = blue,
  filecolor = blue,
  urlcolor = red
}

\makeatletter
\Hy@AtBeginDocument{%
  \def\@pdfborder{0 0 1}
  \def\@pdfborderstyle{/S/U/W 1}
}
\makeatother

\pubyear{0000}
\volume{0}
\firstpage{1}
\lastpage{1}

\begin{document}

\begin{frontmatter}

\title{An Unsupervised Domain-Independent Framework for Automated Detection of Persuasion Tactics in Text}
\runningtitle{An Unsupervised Domain-Independent Framework for Automated Detection of Persuasion Tactics in Text}


\author[A]{\inits{N.N.}\fnms{Rahul Radhakrishnan} \snm{Iyer}\ead[label=e1]{rahuli@andrew.cmu.edu}%
\thanks{Corresponding author. \printead{e1}.}},
\author[B]{\inits{N.}\fnms{Katia} \snm{Sycara}\ead[label=e2]{katia@cs.cmu.edu}},
\runningauthor{Rahul R Iyer et al.}
\address[A]{Language Technologies Institute, \institution{Carnegie Mellon University},
PA, \cny{USA}\printead[presep={\\}]{e1}}
\address[B]{Robotics Institute, \institution{Carnegie Mellon University},
PA, \cny{USA}\printead[presep={\\}]{e2}}

\crefformat{footnote}{\textsuperscript{#2#1#3}}

\begin{abstract}
With the increasing growth of social media, people have started relying heavily on the information shared therein to form opinions and make decisions. While such a reliance is motivation for a variety of parties to promote information, it also makes people vulnerable to exploitation by slander, misinformation, terroristic and predatorial advances. In this work, we aim to understand and detect such attempts at persuasion. Existing works on detecting persuasion in text make use of lexical features for detecting persuasive tactics, without taking advantage of the possible structures inherent in the tactics used. We formulate the task as a multi-class classification problem and propose an unsupervised, domain-independent machine learning framework for detecting the type of persuasion used in text, which exploits the inherent sentence structure present in the different persuasion tactics. Our work shows promising results as compared to existing work.
\end{abstract}


\end{frontmatter}


\section{Introduction}
\label{sec:intro}

Persuasion is being used at every type of forum these days, from politics and military to social media. The amount of information shared on online social media has been growing at unprecedented rates during recent years. The uncontrolled nature of social media makes them vulnerable to exploitation for spreading spam, rumors, slander, and other types of misinformation. An increasing number of people rely, at least in part, on information shared on social media (SM) to form opinions and make choices on issues related to lifestyle, politics, health, and product purchases, and this reliance provides motivation for a variety of parties to promote information.

Detecting persuasion in text helps address many challenging problems: Analyzing chat forums to find grooming attempts of sexual predators; training salespeople and negotiators; and developing automated sales-support systems. Furthermore, ability to detect persuasion tactics in social flows such as SMS and chat forums can enable targeted and relevant advertising. Additionally, persuasion detection is very useful in detecting spam campaigns and promotions on social media: especially those that relate to terrorism. 

Another application is in the military: if it were possible to detect persuasion in conversation, then the military could have information, such as notifications about the enemies trying to influence the local populace. This could result in better intelligence targeting and more focused operations. 

Persuasion identification is also potentially applicable to broader analyses of interaction, such as the discovery of those who shape opinion or the cohesiveness and/or openness of a social group. Existing work on detecting persuasion in text, focuses mainly on lexical features without taking advantage of the inherent structure present in persuasion tactics. 

In this work, we attempt to build an unsupervised, domain-independent model for detecting persuasion tactics, that relies on the sentence structures of the tactics. Our contributions to the literature are: 1) we show that persuasive tactics have inherent sentential structures that can be exploited, 2) we propose an unsupervised approach that does not require annotated data, 3) we propose a way to synthesize prototype strings for the different persuasion tactics, 4) our approach takes much less time to execute as compared to models that require training; for example, our approach is faster than Doc2Vec by a factor of almost $1.5$, 5) our approach is domain-independent, in that it is independent of the vocabulary and can be applied to various domains such as politics, blogs, supreme court arguments etc., with very minimal changes (unlike vector-embedding models and other models that make use of lexical features, because they are dependent on the vocabulary). 

We compare our proposed approach with existing methods that use lexical features, and also some vector embedding models, such as Doc2Vec \cite{le2014distributed}.

We had an intuition about arguments in the persuasive space having similar sentential structures because we had seen and observed a few examples. Consider two examples in the \textit{Reason} category: 1a) \textit{Are we to stoop to their level just because of this argument?}, 1b) \textit{I am angry at myself because I did nothing to prevent this}, and two examples in the \textit{Scarcity} category: 2a) \textit{Their relationship is not something you see everyday}, 2b) \textit{It is only going to go downhill from here}. As we can see, there is a pattern in the structure between arguments in the same category, and there is a structural difference across these two categories. This led us to investigate the problem further and hypothesize our claim.

The rest of the paper is organized as follows: Section \ref{subsec:related} talks about the related work that has been done in the area, section \ref{sec:problem} explains the problem that we are trying to tackle, section \ref{sec:data} gives descriptions about the different datasets that have been used in the paper, section \ref{sec:approach} explains the proposed model and the baselines, section \ref{sec:results} discusses the experimental results obtained, section \ref{sec:applications} goes over some brief applications of the model, and section \ref{sec:discussion} concludes the paper with a discussion and future work.

\subsection{Related Work}
\label{subsec:related}
There has been some work in the literature on detection of persuasion in texts. 
In \cite{young2011microtext}, Young et al. present a corpus for persuasion detection, which is derived from hostage negotiation transcripts. The corpus is called ``NPS Persuasion Corpus'', consists of 37 transcripts from four sets of hostage negotiation transcriptions. Cialdini's model \cite{cialdini2001influence} was used to hand-annotate each utterance in the corpus. There were nine categories of persuasion used: reciprocity, commitment, consistency, liking, authority, social proof, scarcity, other, and non-persuasive. Then algorithms like Naive Bayes, SVM, Maximum Entropy were used for the classification.

Gilbert \cite{gilbert2010persuasion} presented an annotation scheme for a persuasion corpus. A pilot application of this scheme showed some agreement between annotators, but not a very strong one. After revising the annotation scheme, a more extensive study showed significant agreement between annotators. The authors in \cite{ortiz2010machine}, determined that it is possible to automatically detect persuasion in conversations using three traditional machine learning techniques, naive bayes, maximum entropy, and support vector machine. Anand et al. \cite{anand2011believe} describe the development of a corpus of blog posts that are annotated for the presence of attempts to persuade and corresponding tactics employed in persuasive messages. The authors make use of lexical features like unigrams, topic features from LDA, and List count features from the Linguistic Inquiry and Word Count \cite{francis1993linguistic}, and also the tactics themselves, which are provided by human annotators. Tactics represent the type of persuasion being employed: social generalization, threat/promise, moral appeal etc. Carlo et al. \cite{strapparava2010predicting} analyze political speeches and use it in a machine learning framework to classify the transcripts of the political discourses, according to their persuasive power, and predicting the sentences that trigger applause in the audience. In \cite{tan2014effect}, Tan et al. look at the wordings of a tweet to determine its popularity, as opposed to the general notion of author/topic popularity. The computational methods they propose perform better than an average human. In \cite{lukinargument}, Lukin et al. determine the effectiveness of a persuasive argument based on the audience reaction. They report a set of experiments testing at large scale how audience variables interact with argument style to affect the persuasiveness of an argument.

In addition to text, there has been some work on persuasion in the multimedia domain. In \cite{siddiquie2015exploiting}, Siddiquie et al. work on the task of automatically classifying politically persuasive videos, and propose a multi-modal approach for the task. They extract audio, visual and textual features that attempt to capture affect and semantics in the audio-visual content and sentiment in the viewers' comments. They work on each of these modalities separately and show that combining all of them works best. For the experiments, they use \textit{Rallying a Crowd} (RAC) dataset, which consists of over 230 videos from YouTube, comprising over 27 hours of content. Chatterjee et al. \cite{chatterjee2014verbal} aim to detect persuasiveness in videos by analysis of the speaker's verbal behavior, specifically based on his lexical usage and paraverbal markers of hesitation (a speaker’s stuttering or breaking his/her speech with filled pauses, such as \textit{um} and \textit{uh}. Paraverbal markers of hesitation have been found to influence how other people perceive the speaker's persuasiveness. The analysis is performed on a multimedia corpus of 1000 movie review videos annotated for persuasiveness.
Park et al. collected and annotated a corpus of movie review videos in \cite{park2014computational}. From this data, they demonstrate that the verbal and non-verbal behavior of the presenter is predictive of how persuasive they are as well as predictive of the cooperative nature of a dyadic interaction.

Tasks similar to persuasion detection have been explored, such as sentiment detection and perspective detection. Lin et al. investigated the idea of perspective identification at the sentence and document level \cite{lin2006side}. Using the articles from the bitterlemons website\footnote{\url{http://www.bitterlemons.org}}, they were able to discriminate between Palestinian authors and Israeli authors who had written about the same topic. Bikel and Soren used machine learning techniques to differentiate between differing opinions \cite{bikel2007if}. They report an accuracy of $89$\% when distinguishing between 1-star and 5-star consumer reviews, using only lexical features. Recently, several approaches involving machine learning and deep learning have also been used in the visual and language domains \cite{iyer2019event,li2016joint,iyer2016content,iyer2018transparency,li2018object,gupta2016analysis,honke2018photorealistic,iyer2017detecting,radhakrishnan2016multiple,iyer2012optimal,qian2014parallel,iyer2017recomob}.\\

\section{Problem Formulation}
\label{sec:problem}
\subsection{Preliminaries}
Persuasion is an attempt to influence a person's beliefs, attitudes, intentions, motivations, or behaviors. In persuasion, one party (the `persuader') induces a particular kind of mental state in another party (the `persuadee'), like flattery or threats, but unlike expressions of sentiment, persuasion also involves the potential change in the mental state of the other party. Contemporary psychology and communication science further require the persuader to be acting intentionally. Correspondingly, any instance of (successful) persuasion is composed of two events: (a) an attempt by the persuader, which we term the persuasive act, and (b) subsequent uptake by the persuadee. In this work, we consider (a) only, the different persuasive acts, and how to detect them. Working with (b) is a whole other problem. Throughout the rest of the paper, when we say \textit{persuasive arguments}, we mean the former without taking the effectiveness of the persuasion into account. We are only interested in whether the arguments contain persuasion, and if so, the type.

\subsection{Problem Statement}
The main objective of our work is to detect whether a given piece of text contains persuasion or not. If it does, then we can look into the type of persuasion strategy being used, such as threat/promise, outcome, reciprocity etc. In this paper we look at $14$ different persuasion strategies. These are listed in Table \ref{table:tactics}. These are the common tactics for persuasive acts contributed by Marwell and Scmitt \cite{marwell1967dimensions}, Cialdini \cite{cialdini2001influence}, as well as argumentative patterns inspired by Walton et al. \cite{walton2008argumentation}. The intuition behind this investigation along with some examples was discussed in section \ref{sec:intro}.

\begin{figure*}[h]
		\captionsetup{width=0.8\textwidth}
  		\centering
  		\includegraphics[width=\textwidth]{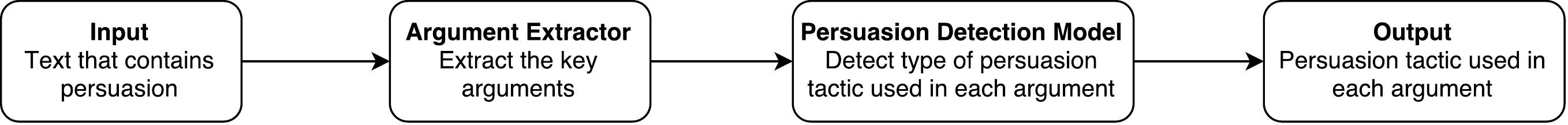}
  		\caption{The outline of the problem considered}
  		\label{fig:problem}
		\end{figure*}

It has to be noted that an entire text is deemed to contain persuasion, if it includes a few arguments that use some of these tactics to persuade. So, it is important to extract such arguments from the text, before applying the persuasion model to them. So, our approach has two steps: 1) a very simple argument extractor model, to extract arguments from a given piece of text, and 2) the output of the extractor is fed into the persuasion detection model, which classifies the arguments into the different tactic classes. This process is represented in the flowchart shown in Figure \ref{fig:problem}. It has to be noted that, in this work, we are not concerned with the effectiveness of the persuasion.

\begin{table*}[h]
\renewcommand{\arraystretch}{1.8} 
\linespread{0.5}\selectfont\centering
\vspace{2ex}
\begin{tabular}{p{2.8cm} | p{2.8cm} | p{2.8cm} | p{2.8cm} | p{2.8cm}}
\toprule
\multicolumn{5}{c}{\textbf{Broad Categories}}\\
\textbf{Outcomes} & \textbf{Generalizations} & \textbf{External} & \textbf{Interpersonal} & \textbf{Other}\\
\midrule
{\small \textbf{Outcome.} {\footnotesize Mentions some particular consequences from uptake or failure to uptake}} & {\small \textbf{Good/Bad Traits.} {\footnotesize Associates the intended mental state with a ``good'' or ``bad'' person’s traits.}} & {\small \textbf{VIP.} {\footnotesize Appeals to authority (bosses, experts, trend-setters)}} & {\small \textbf{Favors/Debts.} {\footnotesize Mentions returning a favor or injury}} & {\small \textbf{Recharacterization.} {\footnotesize Reframes an issue by analogy or metaphor.}}\\
{\small \textbf{Social Esteem.} {\footnotesize States that people the persuadee values will think more highly of them}} & {\small \textbf{Deontic/Moral Appeal.} {\footnotesize Mentions duties or obligations, moral goodness, badness}} & {\small \textbf{Popularity.} {\footnotesize Invokes popular opinion as support for uptake}} & {\small \textbf{Consistency.} {\footnotesize Mentions keeping promises or
commitments}} & {\small \textbf{Reasoning.} {\footnotesize Provides a justification for an argumentative point based upon additional argumentation schemes e.g., causal reasoning, arguments from absurdity}}\\
{\small \textbf{Threat/Promise.} {\footnotesize Poses a direct threat or
promise to the persuadee}} & & & {\small \textbf{Empathy.} {\footnotesize Attempts to make the persuadee connect with someone else’s emotional perspective}} &\\
{\small \textbf{Self-Feeling.} {\footnotesize States that uptake will result in a
better self-valuation by the persuadee}} & & & {\small \textbf{Scarcity.} {\footnotesize Mentions rarity, urgency, or opportunity of some outcome}} &\\
\bottomrule
\end{tabular}
\centering
\caption{List of Persuasion Tactics Used, Taken from \cite{anand2011believe}. We do not consider the broad categories in our experiments, and only work with the finer categories.}
\label{table:tactics}
\end{table*}

\section{Datasets}
\label{sec:data}
We have used datasets from different domains for the experiments to test the robustness of our model. Below are the datasets used for persuasion detection:\\
\indent \textbf{1. ChangeMyView}, an active community on Reddit, provides a platform where users present their own opinions and reasoning, invite others to contest them, and acknowledge when the ensuing discussions change their original views. The training data period is: 2013/01/01 - 2015/05/07, and the test data period is: 2015/05/08 - 2015/09/01. The training dataset contains $3456$ posts and the holdout dataset contains $807$ posts. 
The dataset is organized as follows: each post that is written by a user who wants his views changed, has two argumentative threads -- one that is successful and one that is not. This has been used to determine the persuasion strategies employed by the successful thread. This dataset is used in \cite{tan2016winning}.
	
\textbf{2. Supreme Court Dialogs Corpus}: This corpus contains a collection of conversations from the U.S. Supreme Court Oral Arguments. 1) 51,498 utterances making up 50,389 conversational exchanges, 2) from 204 cases involving 11 Justices and 311 other participants 3) metadata like case-outcome, vote of the Justice, gender annotation etc. This dataset is used in \cite{danescu2012echoes}.
	
 \textbf{3. Blog Authorship Corpora}: This dataset, contributed by Pranav Anand et al. and used in \cite{anand2011believe}, is a subset of the blog authorship corpus. Each directory corresponds to a blog. Each blog has sub-directories corresponding to a day. Inside the day sub-directory, there may be multiple posts; posts are identified with an underscore followed by a number. Out of around 25048 posts, only around 457 were annotated with persuasive acts. Each blog post has been broken down into different ``text tiles'', which are a few sentences long, and each of these tiles are annotated with a persuasive tactic (if present).
	
\textbf{4. Political Speeches}: We collected a number of speeches of Donald Trump and Hillary Clinton, to analyze the kinds of persuasive tactics they use.
	
To train the argumentation extraction model, we use an \textbf{Argumentation Essay Dataset}\footnote{\url{https://www.ukp.tu-darmstadt.de/data/argumentation-mining/argument-annotated-essays-version-2/}}: This consists of about 402 essays. There are two files for each essay - the original essay and the annotated file. Annotations include breaking down the essay into different components: claim, premise, stance. This can be used to train a simple classifier, to identify arguments from text passages.
	
The Blog Authorship Corpus is already annotated, as noted above. In order to have the ground truth, i.e. the annotations, for the other datasets, we needed to annotate the arguments of the corpora with the persuasion tactics mentioned in Table \ref{table:tactics}. For this, we used Amazon Mechanical Turk\footnote{\label{note1}\url{https://www.mturk.com/mturk/}}. Using the argument extraction model, we extracted arguments from all of the corpora combined (excluding the blog authorship corpus)\footnote{\label{note2}The whole dataset, along with the annotation guidelines, classification criteria and the prototype strings (both median and synthetic) for all the persuasion tactics, can be found at \url{https://github.com/rrahul15/Persuasion-Dataset}}. We had each argument annotated by two different turkers, and the turkers were given the freedom to classify a piece of text as either a non-argument or as one of the tactics from Table \ref{table:tactics}. There was about $65$\% inter-annotator agreement, between the turkers and the conflicts were resolved manually. After this, we had a total of $1457$ persuasive arguments from all the datasets combined. The distribution of arguments from the different datasets is given in Table \ref{table:arg-dist}. The guidelines for annotation were built on the ones provided in \cite{anand2011believe}, with some changes\cref{note2}.

\begin{table}[h]
\vspace{2ex}
\begin{tabular}{l | c}
\toprule
\textbf{Dataset} & \textbf{\# Arguments} \\
\midrule
ChangeMyView & 362\\
Supreme Court & 440\\
Political Speeches & 198\\
Blog & 457\\
\bottomrule
\end{tabular}
\centering
\caption{Distribution of arguments from the different datasets}
\label{table:arg-dist}
\end{table}

\section{Technical Approach}
\label{sec:approach}
In this section, we describe our proposed models, along with a couple of baselines for comparison. We describe the baselines in the following section. It has to be kept in mind, as noted in section \ref{sec:data}, that the dataset used consists solely of persuasive arguments.
\subsection{Baselines}
Here, we discuss the different baselines that we use for comparison. We describe a simple supervised approach that makes use of lexical features and then move onto more complicated models involving vector-embedding. In all the supervised approaches, we use a $80:20$ split for training and testing.

\begin{enumerate}
	\item \textbf{Simple Supervised}: Here, the learning phase involved extracting simple textual features from the training set: unigrams, bigrams, without punctuation, and then training an SVM (Support Vector Machine) model, using Sequential Minimal Optimization (SMO) \cite{platt1998sequential}, to learn a model from these features that could be applied to the holdout set. This model was then used to test the remaining posts.

	\item \textbf{Supervised Document Vectors}: This method uses the Doc2Vec model proposed by Quoc and Mikolov \cite{le2014distributed}. First the arguments were separated into different categories based on the persuasion tactic. Then, the Doc2Vec model was applied, to each such cluster, to embed all the arguments into vectors. The prototype vector for each category was then chosen as the mean of all the vectors in that category. To classify the holdout set, one would compute the vector of the argument in consideration and then compute the similarity (cosine) to the prototype vectors. The category which has the highest similarity is the one that is chosen. 
	
	The cosine similarity between two vectors \textbf{a} and \textbf{b} is defined as follows:
	\begin{equation}
		\text{similarity} = \frac{\mathbf{a} \cdot \mathbf{b}}{\lVert \mathbf{a} \rVert \lVert \mathbf{b} \rVert}
	\end{equation}	
	
    \item We also compare our approach with that proposed by \cite{anand2011believe}. Here, the authors make use of different features to account for fewer word-dependent features a) $25$ topic features, which were extracted using Latent Dirichlet Allocation (LDA) \cite{blei2003latent}, with a symmetric Dirichlet prior b) $14$ Tactic count features, i.e., a vector consisting of the count of the tactics. Naive Bayes was used for the classification, to assess the degree to which these feature sets complement each other.
\end{enumerate}
	
\subsection{Proposed Approach}
In this section, we describe the proposed unsupervised, domain-independent approach to identify the persuasion tactics in a given set of arguments. By domain-independence, we mean that our proposed model is robust across different genres, be it political speeches or blogs, and this is important because different domains might have their own vocabulary. Before heading into the details of the algorithm, we present a few useful definitions.

\subsubsection{Preliminaries} In this subsection, we describe a few preliminary concepts\\~\\
\textbf{Parse Tree:} A parse tree is an ordered, rooted tree that represents the syntactic structure of a sentence, according to some context-free grammar. It captures the sentence structure: multiple types of sentences can have a similar sentence structure, even if their vocabularies are not the same. This is the essence of the approach. \\

\noindent \textbf{Edit Distance:} Edit distance is a way of quantifying how dissimilar two strings are to one another by counting the minimum number of operations required to transform one string into the other. Given two strings $a$, and $b$ on an alphabet $\Sigma$, the edit distance $d(a,b)$ is the minimum number of edit operations that transforms $a$ into $b$. The different edit operations are: 1) Insertion of a single symbol, 2) Deletion of a single symbol, 3) Substitution of a single symbol, for another.\\

\noindent \textbf{Median String:} The median string of a set of strings is defined to be that element in the set which has the smallest sum of distances from all the other elements \cite{kohonen1985median}. In our case, the distance between strings is the edit distance. 

\subsubsection{Parse-Tree Model}
\label{subsubsec:parse}
We are proposing a domain-independent classification. The idea is that persuasive arguments may have certain characteristic sentence structures, which we might be able to exploit. The training and testing phase are given below:\\

\textbf{Training Phase}
\begin{enumerate}
    \item As mentioned earlier, we have $14$ different categories for persuasive tactics. We obtain one representative prototype argument for each category. We obtain these in two different ways, which we discuss after detailing the algorithm. 
    \item We then perform phrase-structure parsing on each of these prototype arguments to obtain their parse-trees, which gives the structure of the argument. 
    \item These parse trees are then converted into \textit{parse strings}, keeping the structure intact, and the leaf nodes (the terminal symbols, namely words) are removed, to get a domain independent representation of the structure of the argument. 
    \item By now, we have representative prototype parse strings for each persuasive tactic, i.e. $14$ different prototype parse strings. We use these strings to classify a new argument into one of the persuasive categories. As mentioned earlier, every instance in the dataset is a persuasive argument. This can be construed as the ``training phase''.
\end{enumerate}

\begin{figure*}[h]
		\captionsetup{width=0.8\textwidth}
  		\centering
  		\includegraphics[width=\textwidth]{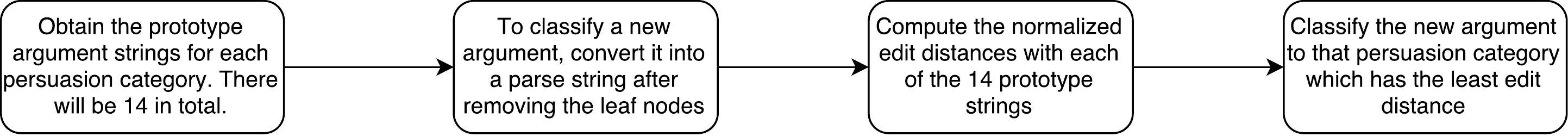}
  		\caption{The parse-tree model explained}
  		\label{fig:parse}
		\end{figure*}

\textbf{Testing Phase}
\begin{enumerate}
    \item For the testing phase, we have to classify a new argument into one of the categories. Since, each new argument in the dataset is persuasive, we don't have to worry about non-arguments. We build a model to account for non-arguments in section \ref{sec:applications}.
    \item Given a new argument, compute its parse string, similar to the procedure used in obtaining the parse strings for the prototype arguments.
    \item To then classify this argument, we compute the normalized edit distances (normalized by the lengths of the strings) between its parse string and the prototype parse strings of each category.
    \item The persuasive category with the least edit distance is logically the most structurally similar to the given argument, and hence the argument is classified into that category.
    \item This process is explained in the flowchart, given in Figure \ref{fig:parse}. 
\end{enumerate}

\begin{figure}[h]
		\captionsetup{width=0.8\textwidth}
  		\centering
  		\includegraphics[scale=0.5]{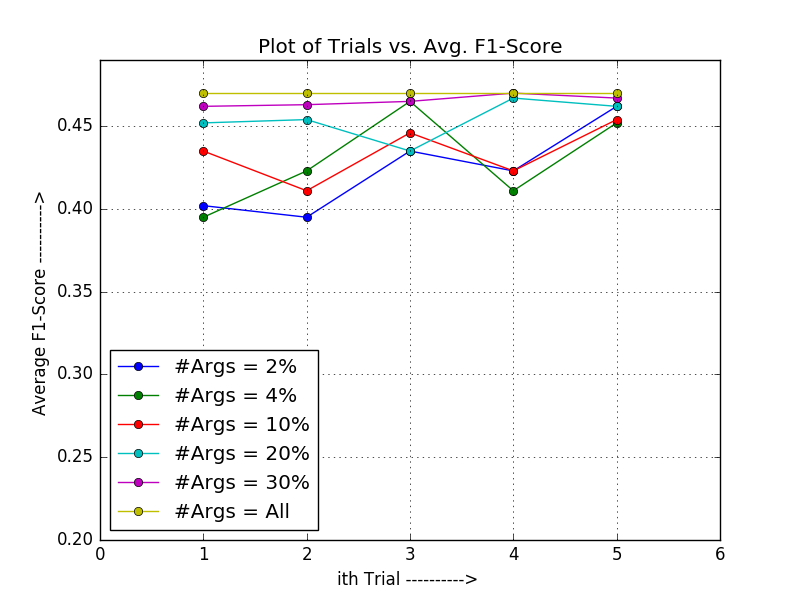}
  		\caption{Trend Graph for Median as the Prototype string}
  		\label{fig:proto1}
		\end{figure}

\begin{figure*}[h]
		\captionsetup{width=0.8\textwidth}
  		\centering
  		\includegraphics[width=\textwidth]{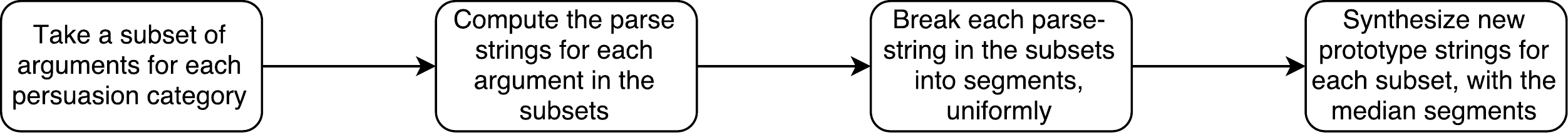}
  		\caption{Outline for Synthesis of the Prototype strings}
  		\label{fig:flowchart-proto}
		\end{figure*}

\begin{figure}[h]
		\captionsetup{width=0.45\textwidth}
  		\centering
  		\includegraphics[scale=0.5]{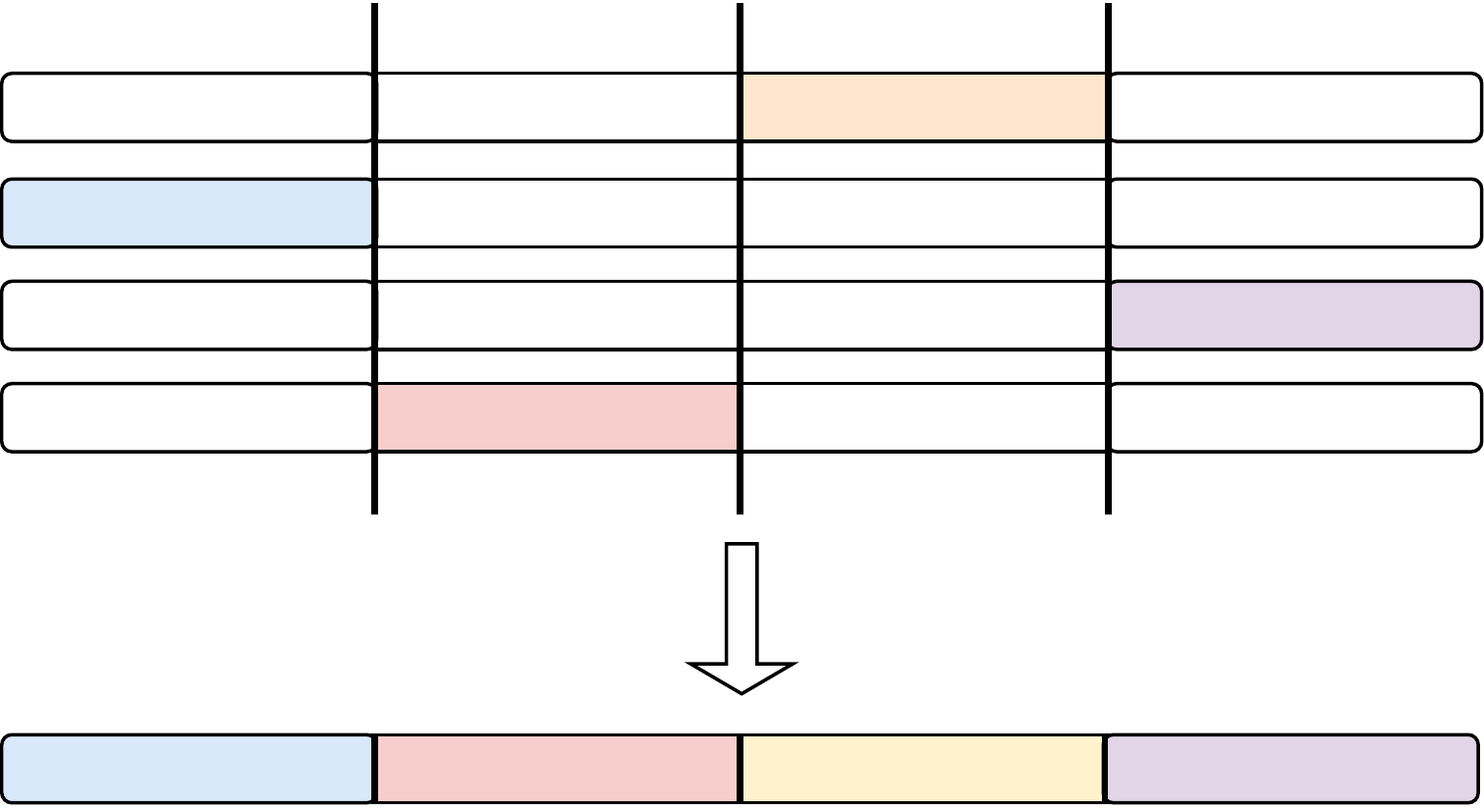}
  		\caption{Synthesis of the prototype string. The coloured segments are the medians of those string segments. In this case, the median of each segment is coming from a different string. This is just for illustrative purposes, it need not always be the case.}
  		\label{fig:prototype_synthesis}
		\end{figure}

\begin{figure}[h]
		\captionsetup{width=0.8\textwidth}
  		\centering
  		\includegraphics[scale=0.5]{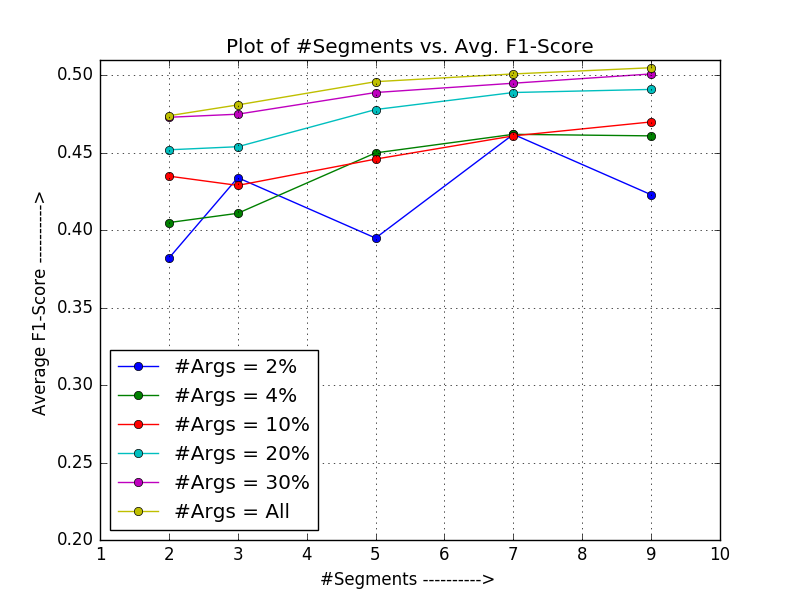}
  		\caption{Trend Graph for Synthetic Prototype string}
  		\label{fig:proto2}
		\end{figure}

\begin{figure*}[h]
        \centering
            \begin{subfigure}{0.4\textwidth}
            \captionsetup{width=0.9\textwidth}
            \centering
            \includegraphics[width=0.9\linewidth]{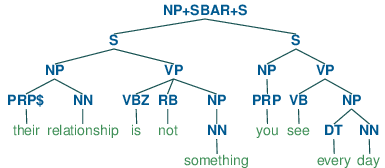}
            \caption{Parse Tree for Scarcity\\Sentence: Their relationship is not something you see everyday\\Parse String: \texttt{(NP+SBAR+S (S (NP (PRP\$) (NN)) (VP (VBZ) (RB) (NP (NN)))) (S (NP (PRP)) (VP (VB) (NP (DT) (NN)))))}}
            \label{subfig:scarcity}
            \end{subfigure}%
            \begin{subfigure}{0.4\textwidth}
            \captionsetup{width=0.9\textwidth}
            \centering
            \includegraphics[width=0.9\linewidth]{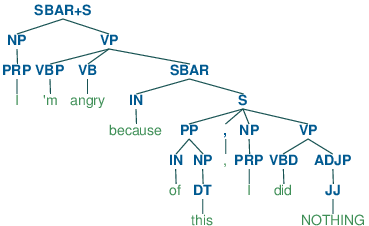}
            \caption{Parse Tree for Reasoning\\Sentence: I'm angry because of this, I did NOTHING\\Parse String: \texttt{(SBAR+S (NP (PRP)) (VP (VBP) (VB) (SBAR (IN) (S (PP (IN) (NP (DT))) (,) (NP (PRP)) (VP (VBD) (ADJP (JJ)))))))}}
            \label{subfig:reason}
            \end{subfigure}
            \caption{Examples of parse trees with the parse-strings for $2$ persuasion tactics: Scarcity and Reason. This is just an illustration to show how the parse strings look like.}
        \label{fig:parse-string}
        \end{figure*}

\textbf{Choosing the Prototype Strings}
We propose two methods to obtain the prototype argument strings. 
\begin{enumerate}
    \item \textit{Median as the Prototype}: Take a set of arguments from each persuasion category and obtain the prototype string for that category as the median string of the set. We now have to determine the ideal size of the set in question. For obvious reasons, we get the best representation if we consider all arguments of that category, but this would require a completely annotated dataset (making the model supervised). 
    
    In order to determine the ideal set size, we conduct additional experiments on a particular dataset, the supreme court dataset, with different parameter values to observe the trend of the performance. We choose different set sizes: 2\%, 5\%, 10\%, 20\%, 30\% and All (the set sizes chosen is a percentage of the total number of arguments in that category). For each set size, we conduct $5$ different trials, choosing a random sample each time, to see the average performance. This trend is shown in Figure \ref{fig:proto1}. As we can see, the performance across successive trials stabilizes as we increase the set size. The performance is best when we consider all the arguments, and is quite stable and close to the best when we consider 30\% of all the arguments. So, we settle for $30\%$ as the ideal set size because we get a stable performance with a very small loss in accuracy and much fewer arguments. Examples of the prototype parse strings, using the median method, for two different persuasion tactics: Reasoning and Scarcity, are given in Figure \ref{fig:parse-string}. We only display prototypes for two of the tactics for purposes of brevity\cref{note2}.
    
    Now, although different arguments in the same category are structurally similar, they may each have certain parts in their structure that capture the essence of that category much better. We rule out taking advantage of these individual segments, when we pick one median argument out of the set. This led us to the second method of obtaining prototype strings.
    
    \item \textit{Synthetic Prototype}: We noted earlier that there could be certain segments in different arguments of the same category, that capture the essence of the category better. To accommodate this, we chop up the different arguments in a set into a number of segments and choose different segments to synthesize an artificial prototype string. To obtain the best $i^{th}$ segment for the synthetic string, we choose the median of the $i^{th}$ segments for all strings in the set. It has to be noted that we chop the strings uniformly. This process of synthesizing the prototype string is illustrated in Figures \ref{fig:flowchart-proto} and \ref{fig:prototype_synthesis}. As before, we need some parameters to tune here. In addition to the optimal set size, we also need to determine the optimal number of segments. 
    
    In order to determine the optimal number of segments and set size, we conduct additional experiments on the supreme court dataset, with different parameter values to observe the trend as before. We choose different set sizes: $2\%$, $5\%$, $10\%$, $20\%$, $30\%$ and All, as before, and different number of segments: 2, 3, 5, 7, and 9. For each (set size, number of segment) pair, we conduct $5$ trails, choosing a random sample for the sets each time, and compute the average performance. This trend is shown in Figure \ref{fig:proto2}. We do not show the performance for different trials as before, rather just the average across the trials. We see that the trend stabilizes as we increase the set size, as before, and the accuracy improves as we consider more number of segments. But here, it is a tradeoff between accuracy and speed because having a large number of segments will require us to compute the median for every segment. We settle for $30\%$ as the ideal set size and $9$ as the ideal number of segments. The prototype strings using this method are not parse strings of meaningful sentences and so we do not display them here.
\end{enumerate}

\subsubsection{Incorporation of Keywords to the Parse-tree Model}
Here, we try to incorporate keywords to the parse-tree model to try to improve the performance.

The basic idea of the algorithm is the same, in that we still form the parse tree, and convert it to a parse string to use in the classification. The only difference is that we don't remove all the leaf nodes

\section{Experimental Results}
\label{sec:results}
In this section, we present the results of our proposed model and compare it with the different approaches described earlier. In the following section, we discuss the metrics that we have used for evaluation.
\subsection{Evaluation Metrics}
The metrics used for evaluation are listed below. 
\begin{enumerate}
\item Precision: The percentage of arguments the system identified as having a particular tactic that in fact had that tactic
	\begin{equation}
	\resizebox{0.435\textwidth}{!}{$
\text{Precision}_t = \frac{|\lbrace\text{retrieved documents}\rbrace| \; \cap \; |\lbrace\text{relevant documents}\rbrace|}{|\lbrace\text{retrieved documents}\rbrace|}
    $}		
	\end{equation}
	where $Precision_t$ is the precision for tactic $t$.
\item Recall: the percentage of arguments of a particular tactic that the system classified into that category
	\begin{equation}
	\resizebox{0.435\textwidth}{!}{$
\text{Recall}_t = \frac{|\lbrace\text{retrieved documents}\rbrace| \; \cap \; |\lbrace\text{relevant documents}\rbrace|}{|\lbrace\text{relevant documents}\rbrace|}
    $}		
	\end{equation}
	where $Recall$ is the recall for tactic $t$.
\item $F_1$-measure: the harmonic mean of precision and recall
	\begin{equation}
		F_{1_t} = \frac{2 \times \text{precision}_t \times \text{recall}_t}{\text{precision}_t + \text{recall}_t}
	\end{equation}
	where $F_{1_t}$ is the $F_1$ measure for tactic $t$.
\end{enumerate} 

It is important to note that, the precision, recall and $F_1$ measure are computed for each persuasion tactic separately, akin to a binary classifier. We report the mean of these measures, over all the tactics, in our experiments.

\subsection{Results}
First, we run the proposed parse-tree model on the arguments extracted from the datasets and obtain the \textit{average per-category accuracy}. The per-category accuracy is defined as the percentage of accurate classifications for a specific category. The categories are the persuasion tactics in our case. For this task, we combined the arguments from each of the $4$ datasets to form a combined set, in order to get an average performance estimate (refer to Table \ref{table:arg-dist} for the distribution of arguments in each dataset). We classified the arguments in the combined set and calculated the fraction of correct classifications for each category. The results are given in Table \ref{table:14}. We do not consider the broad categories in our experiments, and only work with the finer categories.

We also compute the distribution of the tactics in the different datasets. We do this by classifying the arguments in each of the $4$ datasets, and calculating the frequency of appearance of each tactic in the corpus as a percentage over all the arguments in that corpus. These are listed in Tables \ref{table:dist_tactic_cmv}-\ref{table:dist_tactic_hillary}. The ranking of the tactics in these tables, with respect to the percentages, aligns closely with manual evaluations. These distributions are shown just to give an idea of the ranking of the tactics, as predicted by the algorithm (which makes sense intuitively).

\begin{table}[h!]
\vspace{2ex}
\begin{tabular}{l | c}
\toprule
\textbf{Category} & \textbf{Accuracy} \\
\midrule
Reasoning & 79.8\\
Deontic/Moral Appeal & 69.6\\
Outcome & 65.7\\
Empathy & 61.3\\
Threat/Promise & 58.2\\
Popularity & 56.4\\
Recharacterization & 54.9\\
VIP & 53.5\\
Social Esteem & 50.3\\
Consistency & 45.6\\
Favors/Debts & 41.1\\
Self-Feeling & 37.7\\
Good/Bad Traits & 35.5\\
Scarcity & 29.6\\
\bottomrule
\end{tabular}
\centering
\caption{Per-category Accuracy for the parse-tree model, when 14 categories are used}
\label{table:14}
\end{table}

\textbf{ChangeMyView:} Each user posts his/her stance on a particular topic and challenges others to change his opinion. For example, one of the posts was about a man who did not believe in essential-oils and believed that they were destructive, whereas his wife believed the oils were beneficial. He requested the other users to make him change his mind about essential oils by giving him sufficient evidence. If a person is successful in changing the mind of the OP (Original Poster), the OP gives that person a \textit{delta} in their comments. All the conversations are monitored by Reddit and hence the quality is high.

In our dataset, for each post, there are two threads of comments -- one successful in changing the mind of the OP and one that is unsuccessful. We analyzed the persuasive strategies that are used by the successful threads because these would be examples of good uses of the different persuasion tactics. For our purposes of classifying tactics, we could have also used the unsuccessful threads (we are not concerned about the uptake of the persuasion by the persuadee) but we chose not to. Firstly, we extracted the positive comments from the threads (those which were given a delta by the OP). We then applied the parse-tree persuasion model that we developed earlier, to these texts, to perform the classification. Many of the comments had links to other credible sources which listed facts that were opposed to the OP's view. We did not venture into these links. After determining the persuasion strategies used in the comments, we observed that \textit{Reasoning} and \textit{Outcomes} were the most frequently used strategies. A more detailed distribution of tactics is given in Table \ref{table:dist_tactic_cmv}.

\begin{table}[h!]
\vspace{2ex}
\begin{tabular}{l | c}
\toprule
\textbf{Tactic} & \textbf{Percentage}\\
\midrule
Reasoning & 40.7\\
Outcomes & 41.2\\
Good/Bad traits & 10.0\\
Social & 8.1\\
\bottomrule
\end{tabular}
\centering
\caption{Distribution of Persuasion Tactics used in the ChangeMyView dataset.}
\label{table:dist_tactic_cmv}
\end{table}

\textbf{Supreme Court Dataset:} This dataset includes the transcript of the conversation exchanges over 204 cases, along with the outcome of the cases. The outcome could either be \textit{Respondent} or \textit{Petitioner}. Petitioner is the person who files the petition/case against a particular party requesting action on a certain matter, and the respondent is the person against whom the said relief is sought. We have collected all the cases where the petitioner has won and analyzed the argument structure. 

\begin{table}[h!]
\vspace{2ex}
\begin{tabular}{l | c}
\toprule
\textbf{Tactic} & \textbf{Percentage}\\
\midrule
Deontic Appeal & 33.3\\
Reasoning & 35.5\\
Recharacterization & 12.6\\
Outcome & 8.6\\
Empathy & 5.2\\
VIP & 4.8\\
\bottomrule
\end{tabular}
\centering
\caption{Distribution of Persuasion Tactics used in the Supreme Court dataset.}
\label{table:dist_tactic_sc}
\end{table}

We have taken these cases and analyzed the arguments. Using the argumentation-model, we were able to identify the key arguments and then using the parse-tree model, we were able to classify the type of argument that was used. It was found that most of the presented arguments were \textit{Deontic Appeal} and \textit{Reasoning}. The distribution of arguments is given in Table \ref{table:dist_tactic_sc}.

\textbf{Political Speeches:} We analyze the persuasive tactics present in the speeches of political candidates, specifically those of Donald Trump and Hillary Clinton. These distributions are given in Tables \ref{table:dist_tactic_trump} and \ref{table:dist_tactic_hillary}. It was observed that the most frequently used tactic by Trump was Outcome (``Make America Great Again''),
while for Hillary, the most frequently used tactic is Empathy\\

\begin{table}[h!]
\vspace{2ex}
\begin{tabular}{l | c}
\toprule
\textbf{Tactic} & \textbf{Percentage}\\
\midrule
Outcome & 39.1\\
Principles & 31.2\\
VIP & 18.5\\
Reasoning & 11.2\\
\bottomrule
\end{tabular}
\centering
\caption{Distribution of Persuasion Tactics used in Trump's Speeches.}
\label{table:dist_tactic_trump}
\end{table}

\begin{table}[h!]
\vspace{2ex}
\begin{tabular}{l | c}
\toprule
\textbf{Tactic} & \textbf{Percentage}\\
\midrule
Empathy & 35.2\\
Consistency & 33.8\\
Favors/Debts & 18.2\\
Social & 12.8\\
\bottomrule
\end{tabular}
\centering
\caption{Distribution of Persuasion Tactics used in Hillary's Speeches.}
\label{table:dist_tactic_hillary}
\end{table}

Finally, we present the results of the performance of the different algorithms, described earlier, in Table \ref{table:results}. We also performed these experiments in a binary setting: whether a given argument contains persuasion or not. These results are presented in Table \ref{table:binary}. We have run these experiments on the arguments extracted from each dataset (refer to Table \ref{table:arg-dist} for the distribution of arguments in each dataset). The performances are measured by the precision (P), recall (R) and the $F_1$ measure (F), as described earlier.

\begin{table*}[h!]
\vspace{2ex}
\begin{tabular}{|l || c | c | c || c | c | c || c | c | c || c | c | c ||}
\toprule
& \multicolumn{3}{|c||}{Blogs}  & \multicolumn{3}{c||}{ChangeMyView} & \multicolumn{3}{c||}{Supreme Court} & \multicolumn{3}{c||}{Political Speeches}\\
\textbf{Method} & P & R & F & P & R & F & P & R & F & P & R & F\\
\midrule
SVM Baseline & \textbf{0.594} & 0.132 & 0.216 & 0.511 & 0.107 & 0.176 & \textbf{0.605} & 0.051 & 0.094 & 0.454 & 0.038 & 0.070\\\hline
NB+Tactic & 0.361 & \textbf{0.483} & 0.413 & 0.309 & \textbf{0.465} & 0.371 & 0.319 & \textbf{0.511} & 0.393 & 0.267 & \textbf{0.417} & 0.325\\\hline
NB+LDA & 0.098 & 0.132 & 0.112 & 0.071 & 0.116 & 0.088 & 0.083 & 0.151 & 0.107 & 0.032 & 0.045 & 0.037\\\hline
NB+Tactic+LDA & 0.114 & 0.229 & 0.152 & 0.099 & 0.212 & 0.135 & 0.138 & 0.242 & 0.176 & 0.041 & 0.145 & 0.064\\\hline
S-Doc2Vec & 0.493 & 0.439 & 0.464 & 0.472 & 0.427 & 0.448 & 0.496 & 0.442 & 0.467 & 0.411 & 0.392 & 0.401\\\hline
ParseTree & 0.498 & 0.443 & 0.468 & 0.491 & 0.419 & 0.452 & 0.477 & 0.462 & 0.468 & 0.418 & 0.371 & 0.393\\\hline
ParseTree+SP & 0.531 & 0.470 & \textbf{0.498} & \textbf{0.539} & 0.448 & \textbf{0.489} & 0.521 & 0.483 & \textbf{0.501} & \textbf{0.464} & 0.405 & \textbf{0.432}\\
\bottomrule
\end{tabular}
\centering
\caption{Comparison of results for the different methods, considering all the $14$ tactics. Here, ParseTree is the first model proposed, ParseTree+SP stands for the parse-tree model with the synthetic prototype strings, NB stands for Naive Bayes, LDA stands for Latent Dirichlet Allocation, and S-Doc2Vec stands for the supervised version of the Doc2Vec method. NB+Tactic, NB+LDA, NB+Tactic+LDA are the features used by the authors in \cite{anand2011believe}}
\label{table:results}
\end{table*}

\begin{table*}[h!]
\vspace{2ex}
\begin{tabular}{|l || c | c | c || c | c | c || c | c | c || c | c | c ||}
\toprule
& \multicolumn{3}{|c||}{Blogs}  & \multicolumn{3}{c||}{ChangeMyView} & \multicolumn{3}{c||}{Supreme Court} & \multicolumn{3}{c||}{Political Speeches}\\
\textbf{Method} & P & R & F & P & R & F & P & R & F & P & R & F\\
\midrule
SVM Baseline & \textbf{0.741} & 0.179 & 0.288 & \textbf{0.721} & 0.159 & 0.261 & \textbf{0.763} & 0.171 & 0.279 & 0.651 & 0.107 & 0.184\\\hline
NB+Tactic & 0.537 & \textbf{0.672} & 0.597 & 0.515 & \textbf{0.645} & 0.573 & 0.533 & \textbf{0.656} & 0.588 & 0.467 & \textbf{0.599} & 0.525\\\hline
NB+LDA & 0.133 & 0.285 & 0.181 & 0.111 & 0.262 & 0.156 & 0.137 & 0.282 & 0.184 & 0.051 & 0.203 & 0.082\\\hline
NB+Tactic+LDA & 0.169 & 0.437 & 0.244 & 0.144 & 0.416 & 0.214 & 0.154 & 0.441 & 0.228 & 0.097 & 0.354 & 0.152\\\hline
S-Doc2Vec & 0.732 & 0.599 & 0.659 & 0.705 & 0.589 & 0.642 & 0.723 & 0.627 & 0.672 & 0.648 & 0.533 & 0.585\\\hline
ParseTree & 0.701 & 0.603 & 0.648 & 0.692 & 0.589 & 0.636 & 0.716 & 0.614 & 0.661 & 0.633 & 0.539 & 0.582\\\hline
ParseTree+SP & 0.737 & 0.629 & \textbf{0.679} & \textbf{0.721} & 0.623 & \textbf{0.668} & 0.751 & 0.642 & \textbf{0.692} & \textbf{0.661} & 0.562 & \textbf{0.607}\\
\bottomrule
\end{tabular}
\centering
\caption{Comparison of results for the different methods in a \textit{binary} setting. Here, ParseTree is the first model proposed, ParseTree+SP stands for the parse-tree model with the synthetic prototype strings, NB stands for Naive Bayes, LDA stands for Latent Dirichlet Allocation, and S-Doc2Vec stands for the supervised version of the Doc2Vec method. NB+Tactic, NB+LDA, NB+Tactic+LDA are the features used by the authors in \cite{anand2011believe}}
\label{table:binary}
\end{table*}

As can be seen, the domain-independent parse-tree model with synthetic prototype strings performs the best, almost $7-8\%$ better than Doc2Vec. Thus, our intuition that different segments of arguments in the same category capture the essence of the category better than others, is validated. It has to be noted that in a multi-class classification setting, the $F_1$ scores, obtained in Table \ref{table:results}, are reasonable. It also has to be noted that our model performs faster than Doc2Vec, by a factor of almost $1.5$. 

\subsection{Sensitivity Analysis}
We also performed a sensitivity analysis on the parse-tree model to observe its robustness. For this, we combined the arguments from all the $4$ datasets and ran the model on the combined set: 1) using only $10$ instances in the dataset, 2) using only $100$ instances in the dataset, 3) using $1000$ instances in the dataset, and 4) using all the instances in the dataset. The prototype argument strings, for this dataset, were synthesized according to the method mentioned earlier. The results are given in Table \ref{table:sensitivity}. For the cases which did not involve the whole dataset, we randomly sampled $5$ times from the whole corpus and averaged the results. As we can see, the results show that the proposed model is relatively robust and invariant with the amount of data.\\

Another aspect of sensitivity analysis, involving variation of the parameters of the models was discussed earlier in section \ref{subsubsec:parse}.

\begin{table}[h!]
\vspace{2ex}
\begin{tabular}{l | c | c | c}
\toprule
& \multicolumn{3}{c}{Metrics} \\
\textbf{Data} & P & R & F\\
\midrule
10 & 0.370 & 0.375 & 0.372\\
100 & 0.402 & 0.393 & 0.397\\
1000 & 0.491 & 0.387 & 0.432\\
ALL & 0.442 & 0.412 & 0.426\\
\bottomrule
\end{tabular}
\centering
\caption{Sensitivity Analysis of the parse-tree model using $10$, $100$, $1000$ and all instances of data.}
\label{table:sensitivity}
\end{table}

\section{Applications}
\label{sec:applications}
The persuasion-detection model that is proposed here, is very versatile and can be applied in many scenarios. A few applications are detailed below\\

\noindent \textbf{Basic Argument vs. Non-argument Classifier}\\
The parse-tree model proposed can be used as an argument classifier. To test its applicability here, we collected a set of simple sentences\footnote{\url{http://www.cs.pomona.edu/~dkauchak/simplification/}} with less than 10 words. We collected a total of $1000$ such simple sentences. It was our intuition that such simple phrases should have a very low similarity with the different persuasion categories (as they are structurally different). So, we needed to establish a threshold to classify a particular piece of text as an argument versus a non-argument. If the normalized edit distance similarities between the given string and the prototypes of the different categories are less than that threshold (all of them should be less than the threshold), classify it as a non-argument. Else, classify it into its correct persuasion category. Choosing this threshold is not such an easy task for these reasons:\\

\noindent 1) Higher threshold: Lower chance of classifying a non-argument as an argument and Higher chance of classifying an argument as a non-argument\\
2) Lower Threshold: Higher chance of classifying a non-argument as an argument and Lower chance of classifying an argument as a non-argument\\

\begin{figure}[h]
		\captionsetup{width=0.8\textwidth}
  		\centering
  		\includegraphics[scale=0.5]{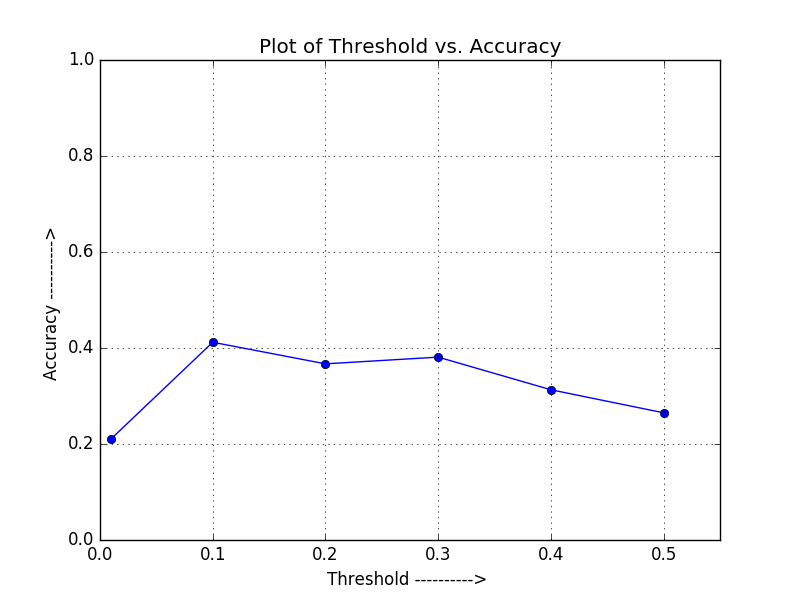}
  		\caption{Graph of Threshold vs. Accuracy ($F_1$ Score)}
  		\label{fig:threshold-acc}
		\end{figure}

So, we needed to choose a threshold that is neither too high nor too low. We tried different thresholds, and we show a graph of threshold vs. accuracy in Figure \ref{fig:threshold-acc}. The accuracy here is the mean $F_1$ score. As we can see, the threshold of $0.1$ seems to work best. For this set threshold, the $F_1$ score is observed to be $0.412$. It can be seen that the performance of this system is not as good as with just persuasive arguments. It has to be noted here that this is not a binary problem: argument vs. non-argument. The $F_1$ score presented here is for the problem of $14$ persuasion tactic categories vs. non-arguments.

This model cannot be used, in this form, as a robust argument classifier yet because some non-arguments have structures similar to some of the persuasion tactics described earlier. For example, consider the sentence: ``The men smoked and most of the women knitted while they talked''. Although this is a non-argument, the model could confuse it with one of the persuasion categories like Reason/Promise. It is for this reason that we consider very simple, straight-forward non-argument sentences of a few words. In order to build a classifier with such capabilities, we would be required to incorporate domain-independent lexical features to the parse tree model (more discussion in section \ref{subsec:future}).\\~\\

\noindent \textbf{Political Speech Analysis}\\
The parse-tree model can also be used to detect spam campaigns on social media, and to detect terrorist campaigns. We analyzed some speeches of Osama Bin Laden to see what kind of tactics he used to influence the people in his speeches. From the analysis, mostly \textit{Empathy} was used. The detailed distribution is given in Table \ref{table:dist_tactic_osama}.

\begin{table}[h!]
\vspace{2ex}
\begin{tabular}{l | c}
\toprule
\textbf{Tactic} & \textbf{Percentage}\\
\midrule
Empathy & 29.7\\
Recharacterization & 20.3\\
Reasoning & 12.9\\
Good/Bad Traits & 20.2\\
Scarcity & 8.0\\
Outcome & 8.9\\
\bottomrule
\end{tabular}
\centering
\caption{Distribution of Persuasion Tactics used in Osama Bin Laden's Speeches.}
\label{table:dist_tactic_osama}
\end{table}

\section{Discussion}
\label{sec:discussion}
We have proposed a fairly simple, domain-independent unsupervised model for detecting types of persuasion used in text. This model can be used in any context/domain because it only uses the inherent structure of the persuasion tactics. This versatility gives it a variety of applications. Almost all persuasive arguments can be classified under the categories mentioned earlier. It has to be noted that the reason we did not include lexical features to our model is because that would make the model slightly domain-dependent. It is for this reason, that we just focused on the structural aspects. Of course, as we mention in section \ref{subsec:future}, it might be possible to include a few lexical terms like \textit{because, if, while} etc., which are domain-independent and tactic-dependent, to further strengthen our model.

From the obtained results, we see that our model's accuracy is highest for the following persuasion tactics (refer to Table \ref{table:14}): reasoning, deontic/moral appeal, outcome, empathy. This is in agreement with the observations made by the authors in \cite{anand2011believe}. This further validates our model. 

From the metrics computed, we see that the performance of the proposed model, with synthesized prototype strings, is better than that of vector-embedding models, such as Doc2Vec, which uses deep learning. There is almost a $7-8\%$ improvement in performance (compared to Doc2Vec), and this over $14$ categories in total is a reasonable amount. This is very interesting because: 1) The Doc2Vec model that we have used as baseline uses all the data, whereas our model just uses a very small subset of the dataset to compute the initial set of $14$ prototype argument strings, one for each category, 2) The Doc2Vec model requires a training phase which can take considerable amount of time, considering its complex structure. In addition, our model runs faster than Doc2Vec by a factor of almost $1.5$, as already mentioned earlier. The reason that our model beats lexical-features based methods could be attributed to the fact that domain-words somehow restrict the performance.

We see that the baseline SVM, on lexical features had considerably high precision but it fell short on recall. That is why we use the $F_1$ measure, as a combination of both aspects of the model: precision and recall. The most suitable model should be the one with a high $F_1$ score, which can be achieved with high values for both precision and recall. We also see that our model performs better, at both binary and multi-class classification, than the approach used in \cite{anand2011believe}.

Sensitivity analysis was also done to test the robustness of the method. As can be seen from the results, the method is fairly stable and robust with respect to the size of the data.

The main takeaway here is that, complex methods like neural networks may not be the best for all tasks. We have shown that with very simple methods like the one we have proposed, we are able to achieve a performance better than the methods discussed in the paper, while avoiding high computational costs and the opacity of results of neural computation based methods.

\subsection{Future Work}
\label{subsec:future}
There is scope for improving the proposed model further. As of now, we just use the sentence structures of the different persuasion tactics. We have not made use of the fact that there could be some domain-independent words for each tactic, like \textit{because, if, while} etc. Incorporation of such keywords into the model could result in improved performance. We will investigate this in the future. Additionally, we will use our approach for different applications, such as detecting spam campaigns, measuring how effective a spam campaign can be (combination of persuasiveness and connectivity in the network, which can be measured by PageRank), identifying terrorist campaigns etc.


\section{Acknowledgements}
This work has been funded by ARO award  \#W911NF-13-1-0416.







\nocite{*} 
\bibliographystyle{ios1}           
\bibliography{sigproc}        

%

\end{document}